\documentclass[runningheads]{llncs}
\usepackage{graphicx}

\usepackage{tikz}
\usepackage{comment}
\usepackage{amsmath,amssymb} 
\usepackage{color}

\usepackage{graphicx}
\usepackage{times}
\usepackage{epsfig}
\usepackage{booktabs}
\usepackage{adjustbox}
\usepackage{multirow}
\usepackage{algorithm}

\usepackage{xcolor}
\usepackage{subcaption}

\usepackage{listings}
\usepackage{xcolor}
\definecolor{codegreen}{rgb}{0,0.6,0}
\definecolor{codegray}{rgb}{0.5,0.5,0.5}
\definecolor{codepurple}{rgb}{0.58,0,0.82}
\definecolor{backcolour}{rgb}{0.95,0.95,0.92}

\lstdefinestyle{mystyle}{
    commentstyle=\color{codegreen},
    keywordstyle=\color{magenta},
    numberstyle=\tiny\color{codegray},
    stringstyle=\color{codepurple},
    basicstyle=\ttfamily\footnotesize,
    breakatwhitespace=false,         
    breaklines=true,                 
    captionpos=b,                    
    keepspaces=true,                 
    numbers=left,                    
    numbersep=5pt,                  
    showspaces=false,                
    showstringspaces=false,
    showtabs=false,                  
    tabsize=2
}

\usepackage{colortbl}
\usepackage[pagebackref,breaklinks,colorlinks]{hyperref}

\usepackage[accsupp]{axessibility}  

\begin{document}
\pagestyle{headings}
\mainmatter
\def\ECCVSubNumber{100}  

\title{Deep Hash Distillation for Image Retrieval} 

\titlerunning{Deep Hash Distillation for Image Retrieval}
%
\author{Young Kyun Jang\inst{1} \and
Geonmo Gu\inst{2} \and
Byungsoo Ko\inst{2} \and
Isaac Kang \inst{1} \and
Nam Ik Cho \inst{1,3}}
\authorrunning{Y. Jang et al.}
%
\institute{ECE \& INMC, Seoul National University, Korea\and 
NAVER Vision \and
IPAI, Seoul National University, Korea \\
\scriptsize{\email{\{kyun0914, korgm403, kobiso62\}@gmail.com}, \email{\{isaackang, nicho\}@snu.ac.kr}}
}


\maketitle

\begin{abstract}
In hash-based image retrieval systems, degraded or transformed inputs usually generate different codes from the original, deteriorating the retrieval accuracy. To mitigate this issue, data augmentation can be applied during training. However, even if augmented samples of an image are similar in real feature space, the quantization can scatter them far away in Hamming space. This results in representation discrepancies that can impede training and degrade performance. In this work, we propose a novel self-distilled hashing scheme to minimize the discrepancy while exploiting the potential of augmented data. By transferring the hash knowledge of the weakly-transformed samples to the strong ones, we make the hash code insensitive to various transformations. We also introduce hash proxy-based similarity learning and binary cross entropy-based quantization loss to provide fine quality hash codes. Ultimately, we construct a deep hashing framework that not only improves the existing deep hashing approaches, but also achieves the state-of-the-art retrieval results. Extensive experiments are conducted and confirm the effectiveness of our work. Code is at  \href{https://github.com/youngkyunJang/Deep-Hash-Distillation}{https://github.com/youngkyunJang/Deep-Hash-Distillation}
\end{abstract}

\section{Introduction}

Especially for retrieval from large-scale databases, {\em hashing} is essential due to its practicality, {\em i.e.,} high search speed and low storage cost. By converting high-dimensional data points into compact binary codes with a hash function, the retrieval system can utilize a simple bit-wise XOR operation to define a distance between the images. A wide variety of works have been studied for learning to hash \cite{Survey_Hash,ITQ,SH,KSH,SDH}, and are still being actively pursued to build fast and accurate retrieval systems.

Recently, techniques for hash learning have been significantly advanced by deep learning, which is called \textit{deep hashing}, and its corresponding works are in the spotlight \cite{Targeted,CB,ExchNet,Active,SelfAdv,AutoTwin,GCNH,Distillhash}. By integrating the hash function into the deep learning framework, the image encoder and hash function are simultaneously learned to generate image hash codes. Regarding the training of deep hashing, the leading techniques are pairwise similarity approaches that use sets of similar or dissimilar image pairs \cite{CNNH,DHN,DCH,HashNet,GPQ,SPQ}, and global similarity in company with classification approaches that use class labels assigned to images \cite{DCBH,CBH,CSQ}.

\begin{figure}[!t]
\centering
\includegraphics[width=0.75\linewidth]{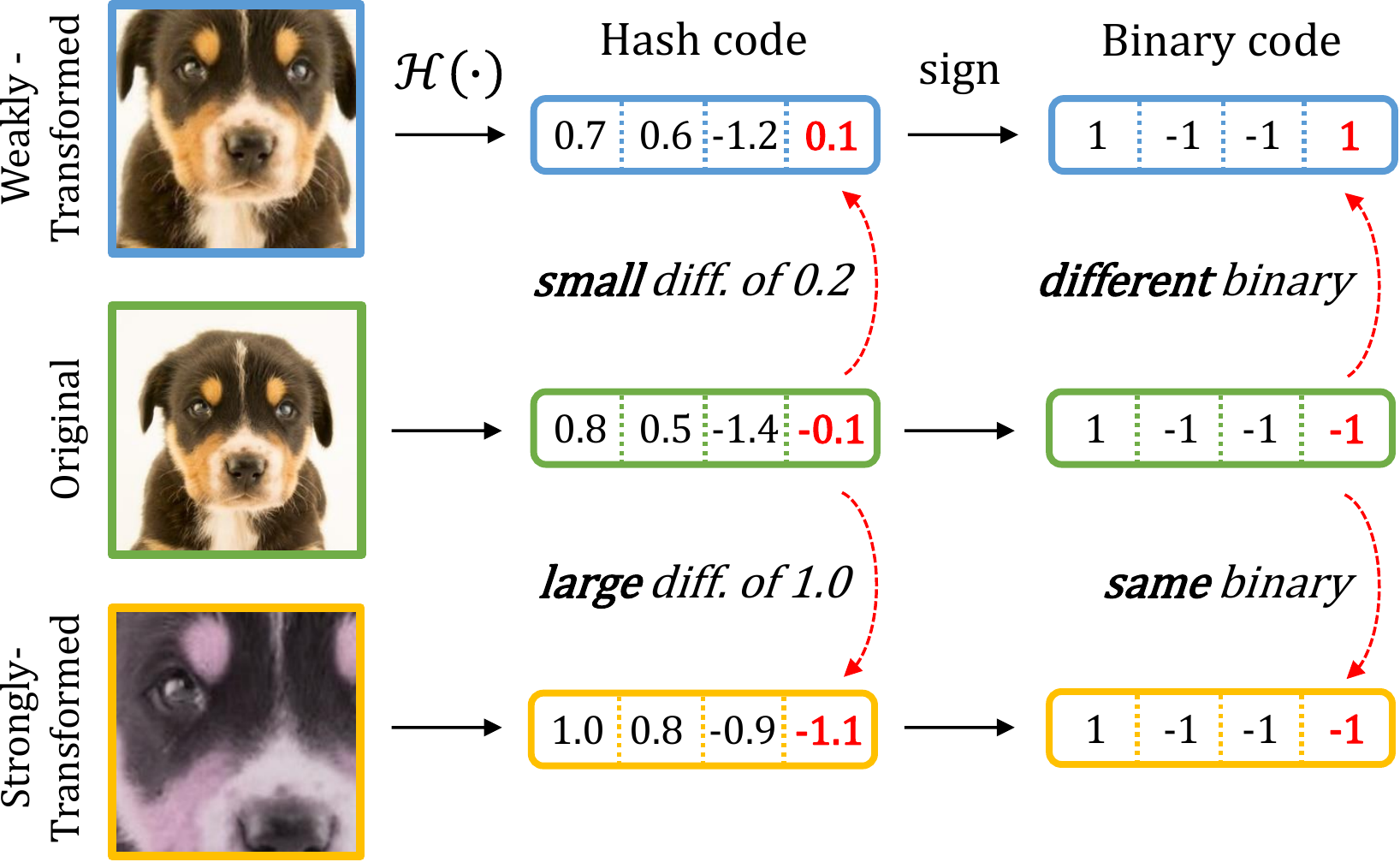}
\caption{Visualization of possible problems in deep hashing due to transformations. The continuous hash code generated by a deep hashing model $\mathcal{H}(\cdot)$ is changed when the input is transformed. In consequence, the binary code quantized with sign operation can also be shifted. However, the degree of transformation cannot be properly reflected in the quantized representation.}
\label{fig:Figure1}
\vspace{-0.5em}
\end{figure}

Since hash-based retrieval systems compute the distance between images with binary codes, corresponding codes need to be quantized with \textit{sign} operation, from the continuous real space to the discrete Hamming space of $\{-1, 1\}$. In this process, the continuously optimized image representation is altered, and quantization error occurs, which in turn degrades the discriminative capability of the hash code. This becomes even more problematic when an input image is transformed and deviated from the original distribution.


To avoid performance degradation due to transformations, the most common solution is to generalize the deep model by training it with augmented data having various transformations. However, it is challenging to apply this augmentation strategy to deep hash training since discrepancy in the representation may occur. Figure~\ref{fig:Figure1} shows an example case that may appear: 1) \textit{The sign of the hash code can be shifted with a slight change}. Specifically, the last element of the weakly-transformed image's hash code differs by 0.2 ($-0.1 \rightarrow 0.1)$ from the original, but it results in $-1 \rightarrow 1$ shift in the Hamming space. 2) \textit{The sign of the quantized hash code does not shift even with the big change in the hash code}. The last element of a strongly transformed image's hash code differs by 1.0 ($-0.1 \rightarrow -1.1$) from the original, resulting in no shifts in the Hamming space. Namely, the use of strong augmentation in deep hashing increases the discrepancy between Hamming and real space, which hinders finding the optimal binary code.

To resolve this issue, we introduce a novel concept dubbed \textit{Self-distilled Hashing}, which customizes self-distillation \cite{RCSD,DeiT,DSD,DINO,FewDistill} to prevent severe discrepancy in deep hash training. Specifically, based on the understanding of the relation between cosine distance and Hamming agreement \cite{KSH,DQN,OneLoss}, we minimize the cosine distance between the hash codes of two different views (transformed results) of an image to maximize the Hamming agreement between their binary outcomes. Further for stable learning, we separate the difficulties of transformations as easy and difficult, and transfer the hash knowledge from easy to difficult, inspired by \cite{SimSiam,DeiT,DINO}.

Moreover, we propose two additional training objectives that optimize hash codes to enhance the self-distilled hashing: 1) a hash proxy-based similarity learning, and 2) a binary cross entropy-based quantization loss. The first term allows the deep hashing model to learn global (inter-class) discriminative hash codes with temperature-scaled cosine similarity. The second term contributes to making the hash code naturally move away from the binary threshold in a classification manner with likelihood estimators.


By combining all of our proposals, we construct a \textbf{D}eep \textbf{H}ash \textbf{D}istillation framework (DHD), which yields discriminative and transformation resilient hash codes for fast image retrieval. We conduct extensive experiments on single and multi-labeled benchmark datasets for image hashing evaluation. In addition, we validate the effectiveness of self-distilled hashing using data augmentation on the existing methods \cite{HashNet,DCH,DPN,CSQ} and show the performance improvements. Furthermore, we establish that DHD is applicable with a variety of deep backbone architectures including vision Transformers \cite{ViT,DeiT,SwinT}. Experimental results verify that self-distilled hashing strategy improves the existing works, and entire DHD framework shows the best retrieval performance.

We can summarize our contributions as follows:

\begin{itemize}
\item To the best of our knowledge, this is the first work to address the discrepancy between real and Hamming space provoked by data augmentation in deep hashing.

\item With the introduction of self-distilled hashing scheme and training loss functions, we successfully embed the power of augmentations into the hash codes.

\item Extensive experiments demonstrate the benefits of our work, improving previous deep hashing methods and achieving the state-of-the-art performances.
\end{itemize}

\section{Related Works}


For a better understanding, we present a brief introduction to the deep hashing methods and the research that inspired our proposal. Refer to a survey~\cite{Survey_Hash} to see details of the early works in non-deep hashing approaches (ITQ \cite{ITQ}, SH \cite{SH}, KSH \cite{KSH}, SDH \cite{SDH}).

\vspace{0.5em}

\noindent \textbf{Deep hashing methods.} Hashing algorithms using deep learning techniques such as Convolutional Neural Network (CNN) are leading the mainstream with striking results. For example, CNNH~\cite{CNNH} utilizes a CNN to generate compact hash codes by training a network with given pairwise label information. DHN~\cite{DHN} learns hash codes by approximating discrete values with relaxation and trains them with supervised signals. HashNet~\cite{HashNet} adopts the inner product to measure pairwise similarity between hash codes and tackles the data imbalance problem by employing weighted maximum likelihood estimation. DCH~\cite{DCH} employs Cauchy distribution to minimize the Hamming distance of the images with the same class label.

\vspace{0.5em}

\noindent \textbf{Hash center-based methods.} There have been several methods to find out class-wise hash representatives (centers), which can provide global similarity to hash codes by including the process of predicting image class labels with hash codes during training \cite{DCBH,CBH,GPQ,CSQ}. CSQ~\cite{CSQ} uses pre-defined orthogonal binary hash targets to guarantee a certain Hamming distance between classes and makes hash codes follow the targets. DPN \cite{DPN} employs randomly assigned target vectors with maximal inter-class similarity and utilizes bit-wise hinge-like loss. Unlike DPN and CSQ, which use a hash target that is not trainable, in our DHD, the hash center is set as a trainable proxy which jointly learns the similarity with the hash codes during training.


\vspace{0.5em}

\noindent \textbf{Self-distillation.} Inspired by knowledge distillation \cite{KD}, self-distillation emerged as a concept that employs a single network to generalize itself in a self-taught fashion, and plenty of works demonstrated its benefits in improving deep model performance \cite{RCSD,DeiT,DSD,DINO}. Many of them utilize a simple Siamese architecture \cite{SimSiam} to explore and learn the visual representation with data augmentation, by contrasting two different augmented results of one image. Similarly, we conduct the self-distillation with augmentations in deep hashing to see the hash codes of two different views of an image simultaneously. Additionally, in accordance with the characteristics of hashing, we consider a method of minimizing the cosine distance that behaves similarly to the distance in the Hamming space to reduce the representation discrepancy during model training. 

\begin{figure*}[!t]
\centering
  \subcaptionbox{Self-distilled hashing. \label{fig:Figure2_a}}{\includegraphics[width=.6\textwidth]{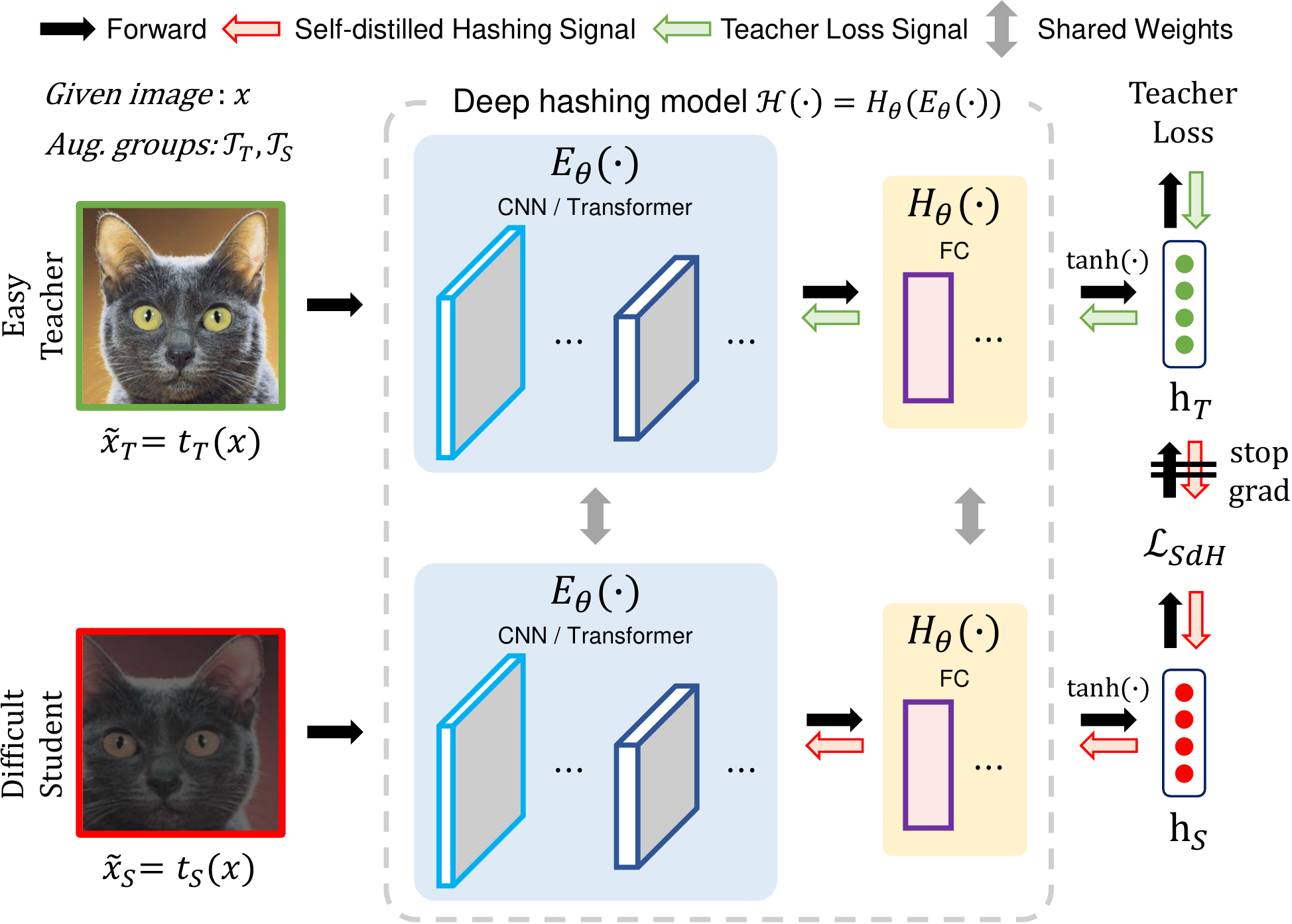}}\hspace{1em}
  \subcaptionbox{Detailed Teacher loss. \label{fig:Figure2_b}}{\includegraphics[width=.31\textwidth]{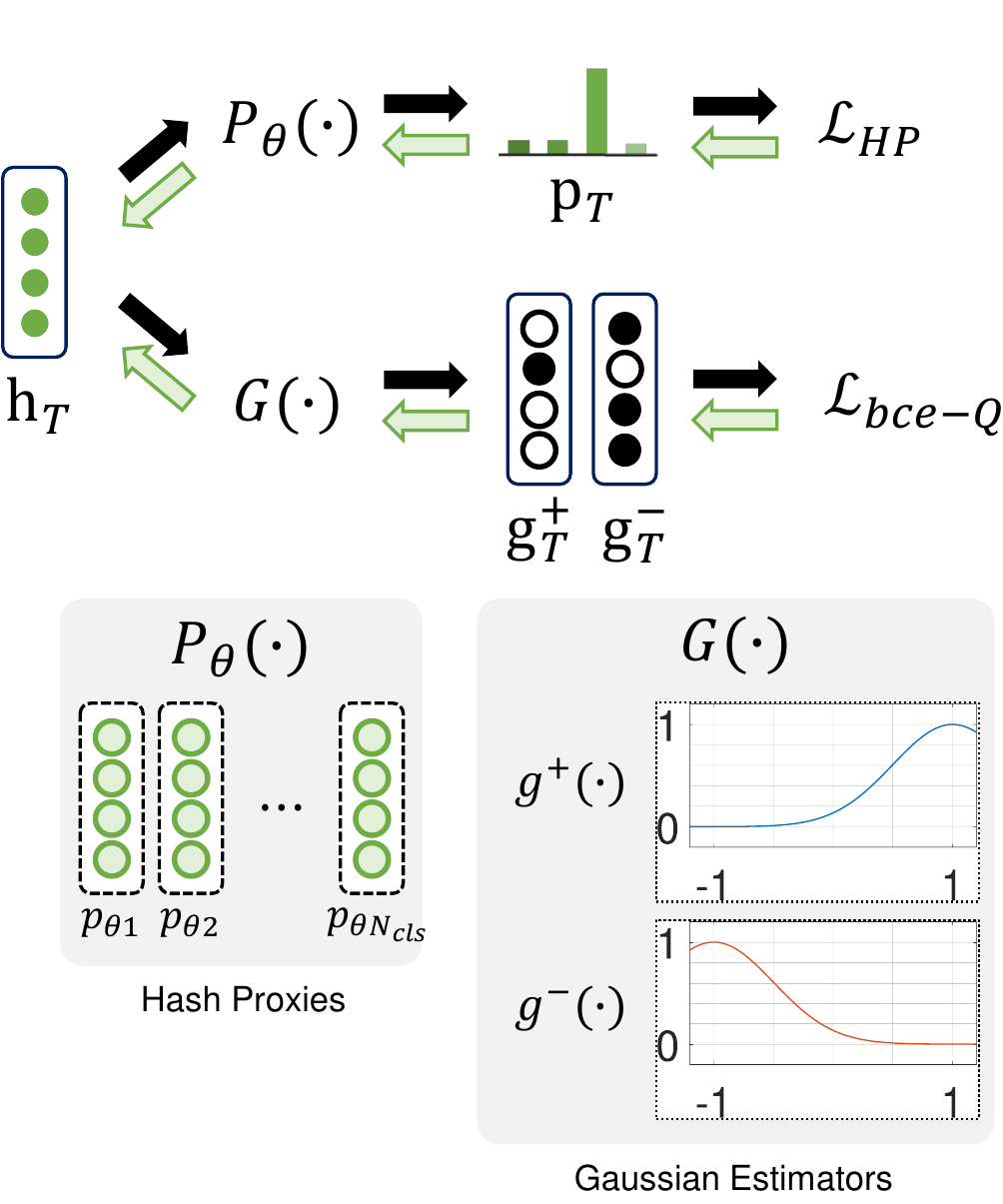}}
  
\caption{The overall training process of Deep Hash Distillation (DHD) framework. (a) From two different augmentation groups, namely Teacher $\mathcal{T}_T$ and Student $\mathcal{T}_S$, randomly sampled transformations ($t_T\sim\mathcal{T}_T$, $t_S\sim\mathcal{T}_S$) are individually applied on the input image $x$ to produce $\tilde{x}_T$ and $\tilde{x}_S$. The deep hashing model $\mathcal{H}(\cdot)$ constructed with the deep encoder $\mathit{E}_\theta(\cdot)$ and the hash function $\mathit{H}_\theta(\cdot)$ of Fully-Connected (FC) layers yields two hash codes $\mathrm{h}_T$ and $\mathrm{h}_S$ which are learned with $\mathcal{L}_{SdH}$. We apply stop gradient operation on $\mathrm{h}_T$ for stable training. (b) Additionally, we employ trainable hash proxies $\mathit{P}_\theta(\cdot)$ which are used to calculate the class-wise prediction $\mathrm{p}_T$ with $\mathrm{h}_T$ to optimize with $\mathcal{L}_{HP}$, and pre-defined Gaussian estimator $\mathit{G}(\cdot)$ to regularize $\mathrm{h}_T$ with $\mathcal{L}_{bce\text{-}Q}$.}
\label{fig:Figure2}
\vspace{-1.0em}

\end{figure*}

\section{Method}


The goal of a deep hashing model $\mathcal{H}$ of Deep Hash Distillation (DHD) is to map an input image $x$ to a $K$-dimensional binary code $\mathrm{b} \in \{-1,1\}^K$ in Hamming space. For this purpose, $\mathcal{H}$ is optimized to find a high quality real-valued hash code $\mathrm{h}$, and then sign operation is utilized to quantize $\mathrm{h}$ as $\mathrm{b}$. Instead of including non-differentiable quantization process in model training, we learn $\mathcal{H}$ in the real space to estimate optimal $\mathrm{b}$ with continuously relaxed $\mathrm{h}$ while fully exploiting the power of data augmentation. We notate trainable components with $\theta$ as a subscript. In the following, $\mathrm{h}$ becomes robust to transformations in \ref{subsection:3.1}, and becomes discriminative and binary-like in \ref{subsection:3.2}.

\subsection{Self-distilled Hashing}
\label{subsection:3.1}

In general, $\mathcal{H}$ is trained in the real space to obtain discriminative $\mathrm{h}$, which should maintain its property in the Hamming space even if quantized to $\mathrm{b}$. Therefore, it is important to align $\mathrm{h}$ and $\mathrm{b}$ to carry a similar representation during training. However, when data augmentation is applied to the training input and the following change occurs in $\mathrm{h}$, there can be misalignment between $\mathrm{h}$ and $\mathrm{b}$, as shown in Figure. \ref{fig:Figure1}. Thus, direct use of augmentations can cause discrepancies in the representation between $\mathrm{h}$ and $\mathrm{b}$, which degrades retrieval performance as we observed in Sec \ref{subsection:4.3}.

\vspace{0.5em}

\noindent \textbf{Hamming distance as cosine distance.} It is noteworthy that the Hamming distance between the binary codes can be interpreted as cosine distance (1-cosine similarity)\footnote{Refer supplementary material for proof.} \cite{KSH,DQN,OneLoss}. That is, the cosine similarity between hash codes $\mathrm{h}_i$ and $\mathrm{h}_j$ can be utilized to approximate the Hamming distance between the binary codes $\mathrm{b}_i$ and $\mathrm{b}_j$ as:

\begin{align}
\mathcal{D}_\mathrm{H}(\mathrm{b}_i,\mathrm{b}_j)\simeq\frac{K}{2}(1-\mathcal{S}\left(\mathrm{h}_i, \mathrm{h}_j)\right)
\label{equation:eqn1}
\end{align}

\noindent where $\mathrm{b}_i=\text{sign}(\mathrm{h}_i)$, $\mathrm{b}_j=\text{sign}(\mathrm{h}_j)$, $\mathcal{D}_\mathrm{H}(\cdot, \cdot)$ denotes Hamming distance, $\mathcal{S}(\cdot,\cdot)$ denotes cosine similarity. That is, the minimized cosine distance between the hash codes minimizes the Hamming distance between the binary codes.

\vspace{0.5em}

\noindent \textbf{Easy-teacher and difficult-student.} As shown in Figure \ref{fig:Figure2_a}, we propose a self-distilled hashing scheme, which supports the training of deep hashing models with augmentations. We employ weight-sharing Siamese structure \cite{SiameseNet} to contrast hash codes of two different views (augmentation results) of an image at once. According to the observations in self-distillation works \cite{SimSiam,DINO}, keeping the output representation of one branch steady has a significant impact on performance gain. Therefore, we configure two separate augmentation groups to provide input views with different difficulties of transformation: one is weakly-transformed easy teacher $\mathcal{T}_T$, and the other is strongly-transformed difficult student $\mathcal{T}_S$. Here, we control the difficulty in a stochastic sampling manner as: employing the same hyper-parameter $s_T$ to all transformations in the group, and make them occur less (weakly) or more (strongly) by scaling their own probability of occurrence. While this manner makes the teacher representation stable, it has the advantage that few extreme examples that produce unstable results are not completely ruled out and contribute to learning. Besides, we stop the gradient of the teacher view's corresponding hash codes to avoid collapsing into trivial solutions \cite{SimSiam,DINO}.

\vspace{0.5em}

\noindent \textbf{Loss computation.} For a given image $x$, self-distillation is conducted with image views as: $\tilde{x}_T=t_T(x)$ and $\tilde{x}_S=t_S(x)$, where $t_T, t_S$ are randomly sampled transformations from $\mathcal{T}_T, \mathcal{T}_S$, respectively. The deep encoder $E_\theta$ and the hash function $H_\theta$ take $\tilde{x}_T$ and $\tilde{x}_S$ as inputs and produce corresponding hash code $\mathrm{h}_T$ and $\mathrm{h}_S$. Then, the proposed Self-distilled Hashing (SdH) loss is computed as:

\vspace{-0.7em}

\begin{align}
\mathcal{L}_{SdH}(\mathrm{h}_T,\mathrm{h}_S)=1-\mathcal{S}(\mathrm{h}_T, \mathrm{h}_S)
\label{equation:eqn2}
\end{align}

\noindent Optimizing $\mathcal{H}$ with $\mathcal{L}_{SdH}$ results in the alignment of $\mathrm{h}_T$ and $\mathrm{h}_S$, and thus $\mathrm{b}_T$ and $\mathrm{b}_S$ as follows Eqn. \ref{equation:eqn1}, which in turn reduces the discrepancy in representation between two differently transformed output binary codes.\footnote{We provide a pseudo-code implementation in supplementary material.}





\vspace{0.5em}

\noindent \textbf{Flexibility.} Note that self-distilled hashing is applicable to the other common deep hashing models \cite {HashNet,DCH,DPN,CSQ} with regard to exploiting data augmentation during training, as shown in Section \ref{subsection:4.3}. Furthermore, various backbones \cite{AlexNet,ResNet,ViT,DeiT,SwinT} can be utilized as deep encoder, and any hash function $H_{\theta}$ configuration is compatible. For simplicity, we employ a single FC layer to obtain a hash code of the desired bits, and apply \textit{tanh} operation at the end to be bound in $[-1, 1]$.

\subsection{For Better Teacher}
\label{subsection:3.2}
Besides self-distilled hashing, additional training signals such as supervised learning loss, and quantization loss are required to obtain the discriminative hash codes. We only employ teacher hash codes to compute the losses, in order to transfer the learned hash knowledge to the student's codes. 

\vspace{0.5em}

\noindent \textbf{Proxy-based similarity learning.} Supervised hash similarity learning with pre-defined orthogonal binary hash targets has shown great performance \cite{CSQ,DPN,OneLoss}. However, the hash target has limitations in that 1) it requires a complex initialization process, and 2) it allocates the same Hamming distance between centers so detailed distances according to semantic similarity cannot be learned. Therefore, as shown in Figure \ref{fig:Figure2_b}, we introduce a proxy-based representation learning \cite{ProxyNCA,Closer,Deform} in deep hashing by using a collection of trainable hash proxies $P_\theta$. It has the advantage that the proxies are simply initialized with randomness, and being able to learn semantic similarity into the proxies. In terms of training, we first use $P_\theta$ to compute class-wise prediction $\mathrm{p}_T$ with $\mathrm{h}_T$ as:

\begin{align}
\mathrm{p}_T=\left[\mathcal{S}(p_{\theta1},\mathrm{h}_T),\mathcal{S}(p_{\theta2},\mathrm{h}_T), ..., \mathcal{S}(p_{\theta N_{cls}},\mathrm{h}_T)\right]
\label{equation:eqn3}
\end{align}

\noindent where $p_{\theta_i}$ is a hash proxy assigned to each of the $i$-th class and $N_{cls}$ denotes the number of classes to be distinguished. Then, we use $\mathrm{p}_T$ to learn the similarity with class label $\mathrm{y}$ by computing Hash Proxy (HP) loss as:

\begin{align}
\mathcal{L}_{HP}(\mathrm{y}, \mathrm{p}_T, \tau)=H\left(\mathrm{y}, \text{Softmax}(\mathrm{p}_T/\tau)\right)
\label{equation:eqn4}
\end{align}

\noindent where $\tau$ is a temperature scaling hyper-parameter, $H(\mathrm{u},\mathrm{v})=-\sum\nolimits_{k} u_k\log v_k$ is a cross entropy, and Softmax operation is applied along the dimension of $\mathrm{p}_T$. Note that, similar to Eqn \ref{equation:eqn1}, $\mathcal{L}_{HP}$ is designed to learn Hamming agreement with temperature scaling.


\vspace{0.5em}

\noindent \textbf{Reducing quantization error.} To make continuous hash code elements act like binary bits, the deep hashing methods \cite{DCH,HashNet,DCBH,CSQ} aim to reduce the quantization error by minimizing the distance (e.g. Euclidean) between the hash code bit and its closest binary goal ($+1$ or $-1$) in a regression manner. However, since the purpose of hashing is to classify the sign of each bit, it is a more natural choice to view it as a binary classification: maximum likelihood problem. Hence, we adopt a pre-defined Gaussian distribution estimator $g(h)$ of mean $m$ and standard deviation $\sigma$ as:

\begin{align}
g(h)=\exp\left(-\frac{(h-m)^2}{2\sigma^2}\right)
\label{equation:eqn5}
\end{align}

\noindent to evaluate the binary likelihood of hash code element $h$. By employing two estimators: $g^{+}$ of $m=1$, and $g^{-}$ of $m=-1$ with the same $\sigma$, we compute the likelihoods and a Binary Cross Entropy-based (BCE) quantization loss as:

\begin{align}
\mathcal{L}_{bce\text{-}Q}(\mathrm{h}_T)=\frac{1}{K}\sum_{k=1}^{K}\left(H_b\left(b_k^{+},g^{+}_{k}\right)+H_b\left(b_k^{-},g^{-}_{k}\right)\right)
\label{equation:eqn6}
\end{align}

\noindent where $H_b(u,v)=-u\log v + (1-u)\log (1-v)$ is a binary cross entropy, $g_k^{+}$, $g_k^{-}$ denotes $k$-th hash code element's estimated likelihood: $g_k^{+}=g^{+}(h_k)$, $g_k^{-}=g^{-}(h_k)$, and $b_k^{+}$, $b_k^{-}$ denotes binary likelihood labels which are obtained (refer Figure \ref{fig:Figure2_b}) as: 

\begin{align}
b_k^{+} = \frac{1}{2}\left(\text{sign}(h_k)+1\right), b_k^{-} = 1-b_k^{+}
\label{equation:eqn7}
\end{align}

\noindent As a result, quantization error is reduced by a binary classification loss with the given estimators, allowing to use the merits of cross entropy presented in~\cite{BCE}. Note that, $\mathcal{L}_{bce\text{-}Q}$ is also applied to hash proxies to make them act as continuously relaxed binary codes.
 
\subsection{Training}



\noindent \textbf{Total training loss.} Suppose we are given a training mini-batch of $N_B$ data points: $\mathcal{X}_B=\{(x_1, \mathrm{y}_1),...,(x_{N_B}, \mathrm{y}_{N_B})\}$ where each image $x_i$ is assigned a label $\mathrm{y}_i\in \{0,1\}^{N_{cls}}$. Training views are obtained as $\tilde{x}_{Ti}=t_{Ti}(x_i)$ and $\tilde{x}_{Si}=t_{Si}(x_i)$ for all data points, where $t_{Ti}\sim\mathcal{T}_T$ and $t_{Si}\sim\mathcal{T}_S$. Total loss $\mathcal{L}_{T}$ for DHD is computed with $\mathcal{X}_B$ as:

\begin{align}
\mathcal{L}_{T}(\mathcal{X}_B)=\frac{1}{N_B}\sum^{N_B}_{n=1}\left(\mathcal{L}_{HP} + \lambda_1\mathcal{L}_{SdH} + \lambda_2\mathcal{L}_{bce\text{-}Q}\right)
\label{equation:eqn8}
\end{align}

\noindent where $\lambda_1$ and $\lambda_2$ are hyper-parameters that balance the influence of the training objectives. The entire DHD framework is trained in an end-to-end fashion.

\vspace{0.5em}

\noindent \textbf{Multi-label case.} In the case of determining semantic similarity between multi-hot labeled images, the previous works \cite{CNNH,DHN,CSQ} simply checked whether the images share at least one positive label or not. However, learning with the above similarity has limitations in that the label dependency \cite{Multilabel} is ignored. Thus, we aim to capture the intelligence that appears in label dependency by utilizing the Softmax cross entropy with the normalized multi-hot label $\mathrm{y}$. Specifically, $\mathrm{y}$ is converted as $\mathrm{y}=\mathrm{y}/\lVert\mathrm{y}\rVert_1$ to balance the contribution of each label, and the same $\mathcal{L}_{HP}$ is computed to optimize the deep hashing model for multi-label image retrieval.

\section{Experiments}

\subsection{Setup}

\noindent \textbf{Datasets.} To evaluate our DHD, we conduct experiments against several conventional and modern methods. Three most popular hashing based retrieval benchmark datasets are explored\footnote{The details of each dataset are described in the supplementary material.}, and we explain the composition of each dataset in Table \ref{table:Table1}.

\vspace{-1.0em}

\begin{table}[h]
\centering
\caption{Description of the image retrieval datasets.}
\begin{adjustbox}{width=0.48\textwidth}
\begin{tabular}{ccccc}
\toprule
Dataset       & $\#$ Database   & $\#$ Train & $\#$ Query & $N_{c}$ \\ \midrule
ImageNet \cite{ImageNet}      & 128,503 & 13,000   & 5,000    & 100        \\ \midrule
NUS-WIDE \cite{Nus-Wide}      & 149,736 & 10,500   & 2,100    & 21         \\ \midrule
MS COCO \cite{MSCOCO}       & 117,218 & 10,000   & 5,000    & 80         \\ 
\bottomrule
\end{tabular}
\end{adjustbox}
\label{table:Table1}
\end{table}

\noindent \textbf{Evaluation metrics.} We follow the protocol utilized in deep hashing \cite{HashNet,DCH,CSQ} to evaluate our approach on both single-labeled and multi-labeled datasets. Specifically, we employ three metrics: 1) mean average precision (\textbf{mAP}), 2) precision-recall curves (\textbf{PR curves}), and 3) precision with respect to top-$M$ returned image (\textbf{P@Top-$M$}). Regarding mAP score computation, we select the top-$M$ images from the retrieval ranked-list results. The returned images and the query image are considered relevant whether one or more class labels are the same. We set binary code length: hash code dimensionality $K$ as 16, 32, and 64, to examine the performance according to the code size.

\begin{table*}[!t]
\centering
\caption{mean Average Precision (mAP) scores for different bits on three benchmarks.}
\begin{adjustbox}{width=0.87\textwidth}
\begin{tabular}{c|c|c|c|c|c|c|c|c|c|c}
\toprule
\multirow{2}{*}{Method}    & \multirow{2}{*}{Backbone} & \multicolumn{3}{c|}{ImageNet} & \multicolumn{3}{c|}{NUS-WIDE} & \multicolumn{3}{c}{MS COCO} \\ \cmidrule{3-11} 
                           &                           & 16-bit   & 32-bit  & 64-bit  & 16-bit   & 32-bit  & 64-bit  & 16-bit  & 32-bit  & 64-bit  \\ \midrule
ITQ \cite{ITQ}                        & \multirow{4}{*}{Non-deep} & 0.266    & 0.436   & 0.576   & 0.435    & 0.396   & 0.365   & \textbf{0.566}   & 0.562   & 0.502   \\
SH \cite{SH}                        &                           & 0.210    & 0.329   & 0.418   & 0.401    & 0.421   & 0.423   & 0.495   & 0.507   & 0.510   \\
KSH \cite{SH}                       &                           & 0.160    & 0.298   & 0.394   & 0.394    & 0.407   & 0.399   & 0.521   & 0.534   & 0.536   \\
SDH \cite{SDH}                       &                           & \textbf{0.299}    & \textbf{0.455}   & \textbf{0.585}   & \textbf{0.575}    & \textbf{0.590}   & \textbf{0.613}   & 0.554   & \textbf{0.564}   & \textbf{0.580}   \\ \midrule
CNNH \cite{CNNH}                      & \multirow{6}{*}{AlexNet \cite{AlexNet}}  &  0.315    & 0.473   & 0.596   & 0.655    & 0.659   & 0.647   & 0.599   & 0.617   & 0.620  \\
DNNH \cite{DNNH}                      &                           & 0.353    & 0.522   & 0.610   & 0.703    & 0.738   & 0.754   & 0.644   & 0.651   & 0.647   \\
DHN \cite{DHN}                       &                           & 0.367    & 0.522   & 0.627   & 0.712    & 0.739   & 0.751   & 0.701   & 0.710   & 0.735 \\
HashNet \cite{HashNet}                   &                           & 0.425    & 0.559   & 0.649   & 0.720    & 0.745   & 0.758   & 0.685   & 0.714   & 0.742 \\
DCH \cite{DCH}                      &                           &  0.636    & 0.645   & 0.656   & 0.740    & 0.752   & 0.763   & 0.695   & 0.721   & 0.748  \\
\rowcolor{cyan!10} DHD(Ours)                  &                           &  \textbf{0.657}    & \textbf{0.701}   & \textbf{0.721}   & \textbf{0.780}    & \textbf{0.805}   & \textbf{0.820}   & \textbf{0.749}   & \textbf{0.781}   & \textbf{0.792}  \\ \midrule
DPN \cite{DPN}                       & \multirow{3}{*}{ResNet \cite{ResNet}}   &  0.828    & 0.863   & 0.872   & 0.783    & 0.816   & 0.838   & 0.796   & 0.838   & 0.861  \\
CSQ \cite{CSQ}                       &                           & 0.851    & 0.865   & 0.873   & 0.810    & 0.825   & 0.839   & 0.750  & 0.824   & 0.852  \\\rowcolor{cyan!10}
DHD(Ours)                  &                           &  \textbf{0.864}    & \textbf{0.891}   & \textbf{0.901}   & \textbf{0.820}    & \textbf{0.839}   & \textbf{0.850}   & \textbf{0.839}   & \textbf{0.873}   & \textbf{0.889}  \\ \midrule 
\rowcolor{cyan!10}  & ViT \cite{ViT}                      & 0.927    & 0.938   & 0.944   & 0.837    & 0.862   & 0.870   & 0.886   & 0.919   & 0.939   \\
                         \rowcolor{cyan!10} & DeiT \cite{DeiT}                     & 0.932    & 0.943   & 0.948   & 0.839    & 0.861   & 0.867   & 0.883   & 0.913   & 0.925  \\ 
                      \rowcolor{cyan!10}  \multirow{-3}{*}{DHD(Ours)}  & SwinT \cite{SwinT}                    & \textbf{0.944}    & \textbf{0.955}   & \textbf{0.956}   & \textbf{0.848}    & \textbf{0.867}   & \textbf{0.875}   & \textbf{0.894}   & \textbf{0.930}   & \textbf{0.945}  \\
\bottomrule
\end{tabular}
\end{adjustbox}
\label{table:Table2}
\end{table*}








\subsection{Implementation Details}

\noindent \textbf{Data augmentation.} Following the works presented in \cite{Simclr}, we choose family $\mathcal{T}$ of five image transformations: 1) resized crop, 2) horizontal flip, 3) color jitter, 4) grayscale, and 5) blur, where all of each are sampled uniformly with a given probability and sequentially applied to the inputs. We keep the internal parameters of each transformation equal to \cite{Simclr}. For self-distilled hashing, we configure two groups with $\mathcal{T}$, where the difficult student group is $\mathcal{T}_S=\mathcal{T}$, and the easy teacher group $\mathcal{T}_T$ is configured by scaling all transform occurrence with $s_T$, which is in the range of $(0, 1]$. We set $\mathcal{T}_T$ as the default for the methods trained without SdH.

\vspace{0.5em}

\noindent \textbf{Experiments.} Retrieval experiments are conducted by dividing backbones as: Non-deep, AlexNet \cite{AlexNet}, ResNet (ResNet50) \cite{ResNet} , and vision Transformers \cite{ViT,DeiT,SwinT}. For non-deep hashing approaches: ITQ \cite{ITQ}, SH \cite{SH}, KSH \cite{KSH} and SDH \cite{SDH}, we report the results directly from the latest works \cite{HashNet,DCH,CSQ} for comparison. We set up the same training environment by leveraging PyTorch framework and the image transformation functions of kornia \cite{KORNIA} library for augmentation. We employ Adam optimizer \cite{Adam} and decay the learning rate with cosine scheduling \cite{CosLR} for training deep hashing methods. Especially for DHD hyper-parameters\footnote{More details can be found in the supplementary material.}, $s_T$ is set to 0.2 for AlexNet, and 0.5 for other backbones. $\tau$ is set by considering $N_{cls}$ as $\{0.2, 0.6, 0.4\}$ for $\{$ImageNet, NUS-WIDE, MS COCO$\}$, respectively. $\lambda_1$ and $\lambda_2$ are set equal to 0.1 for a balanced contributions each training objective, and $\sigma$ in $\mathcal{L}_{bce\text{-}Q}$ is set to 0.5 as default.






\subsection{Results and Analysis}
\label{subsection:4.3}

\noindent \textbf{Comparison with others.} The mAP scores are calculated by varying the top-$M$ for each dataset as: ImageNet@1000, NUS-WIDE@5000 and MS COCO@5000 to make a fair comparison with previous works \cite{HashNet,DCH,CSQ}. The results are listed in Table \ref{table:Table2}, where the highest score for each backbone is shown in bold, and we \colorbox{cyan!10}{highlight} our DHD method. Among the non-deep hashing methods, SDH shows the best retrieval results by employing supervised label signals in hash function learning. Deep hashing methods generally outperform non-deep hashing ones, since elaborately labeled annotations are fully utilized during training. For ImageNet, NUS-WIDE, and MS COCO, averaging the mAP scores of all bit lengths yields 36.3\%p, 33.7\%p, and 25.0\%p differences between the non-deep and deep methods, respectively.

Notably, our DHD shows the best mAP scores for all datasets in every bit length with every deep backbone architecture. In particular for AlexNet backbone hashing approaches, DHD shows performance improvement of 16.3\%p, 7.9\%p, and 9.2\%p by averaging the mAP scores of all bit lengths in three dataset results orderly, compared to others. To make a comparison with ResNet backbone methods, DHD also achieves 2.7\%p, 1.8\%p, and 4.7\%p higher retrieval scores on average. In line with the recent trend of other computer vision tasks, we \textit{first} introduce Transformer-based image representation learning architectures: ViT \cite{ViT}, DeiT \cite{DeiT}, and SwinT \cite{SwinT} to the hashing community and perform retrieval experiments. As reported, when the Transformer is integrated into the DHD framework, it delivers outstanding results for the benchmark image datasets with the increase of 5.8\%p, 2.2\%p, and 4.8\%p, in the same as above, compared to ResNet backbone DHD.

\begin{figure*}[!t]
\centering
\subcaptionbox{ImageNet \label{fig:Figure3_a}}{\includegraphics[height=2.9cm]{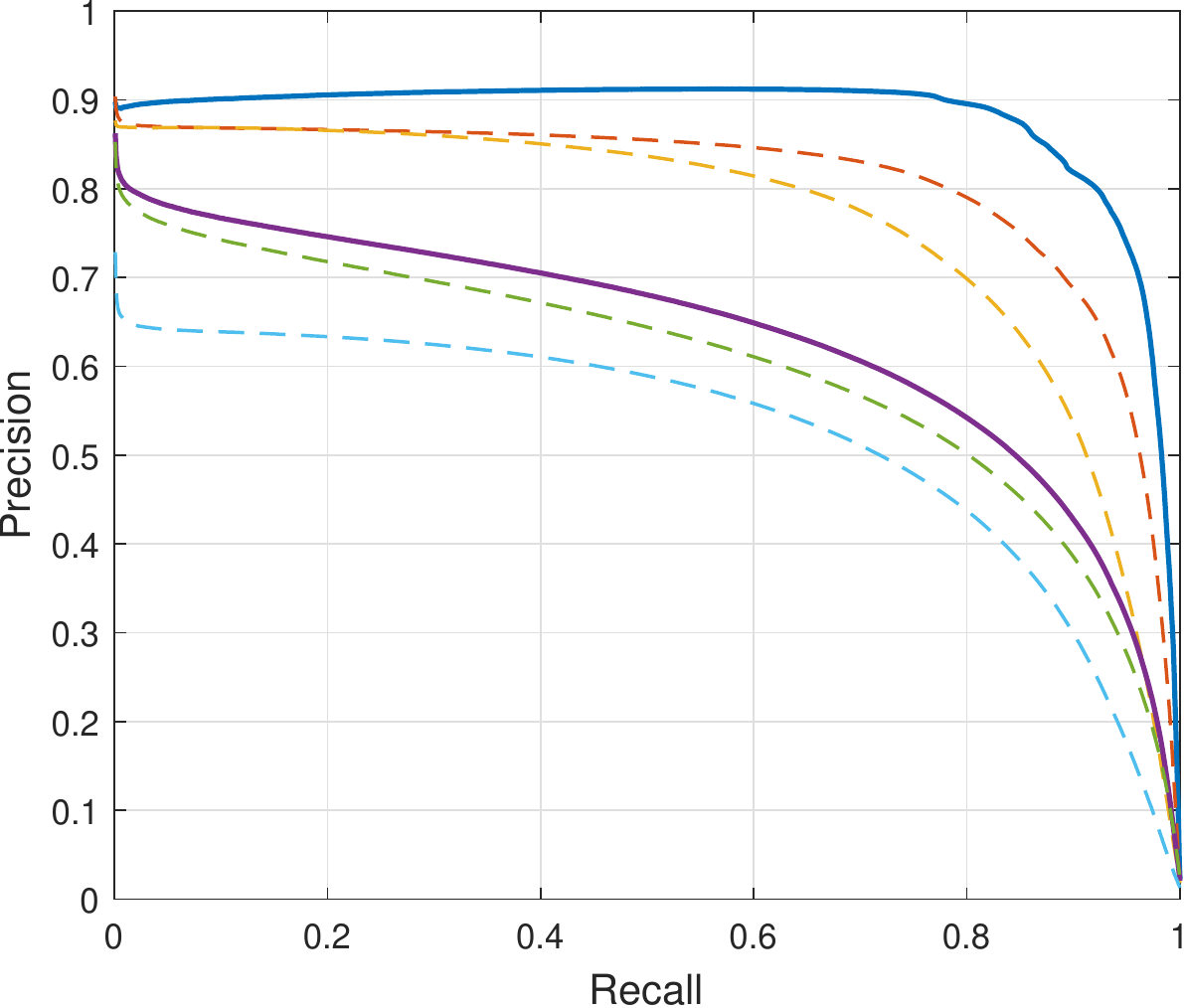}}
\subcaptionbox{NUS-WIDE \label{fig:Figure3_b}}{\includegraphics[height=2.9cm]{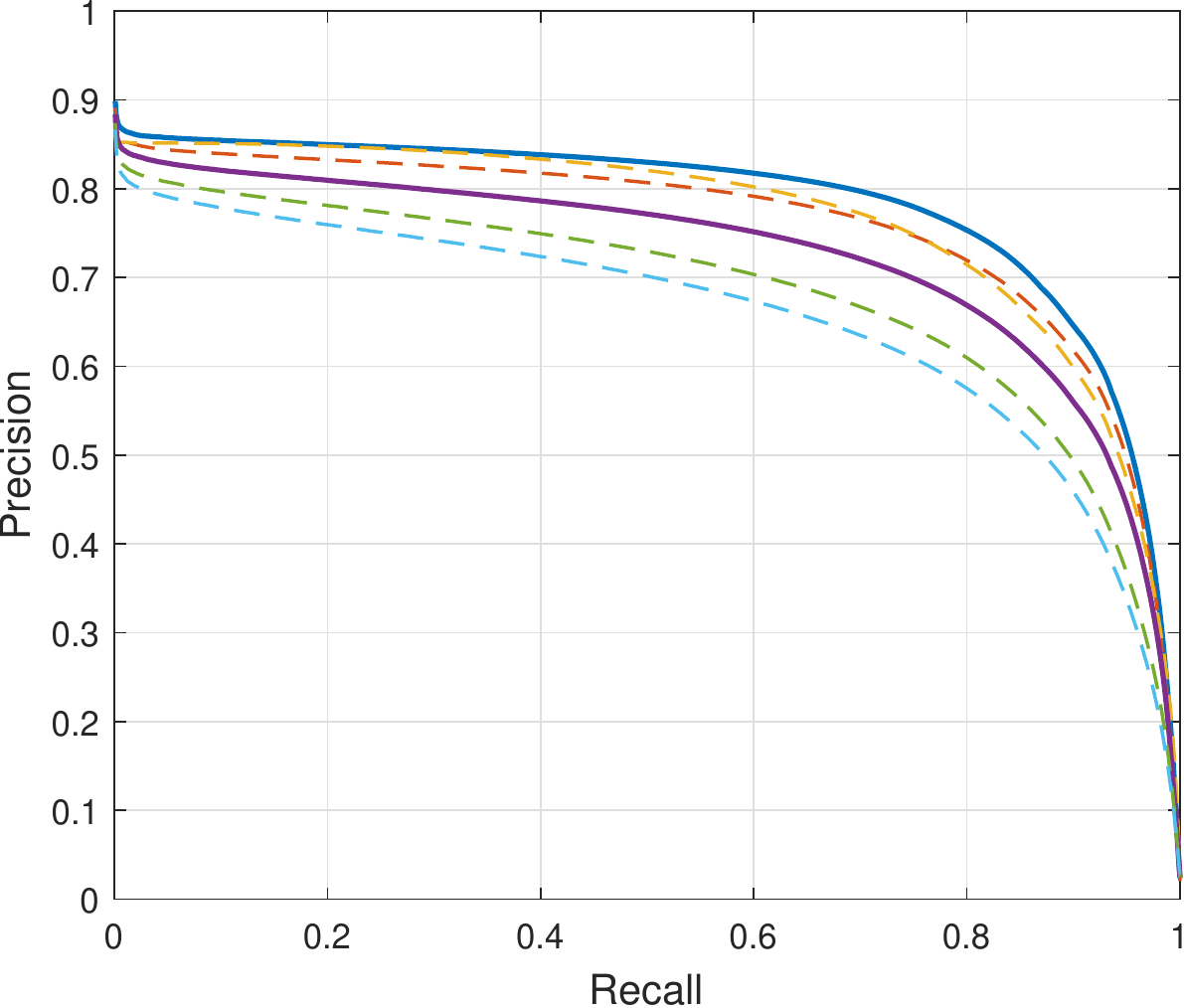}}
\subcaptionbox{MS COCO \qquad\qquad \label{fig:Figure3_c}}{\includegraphics[height=2.9cm]{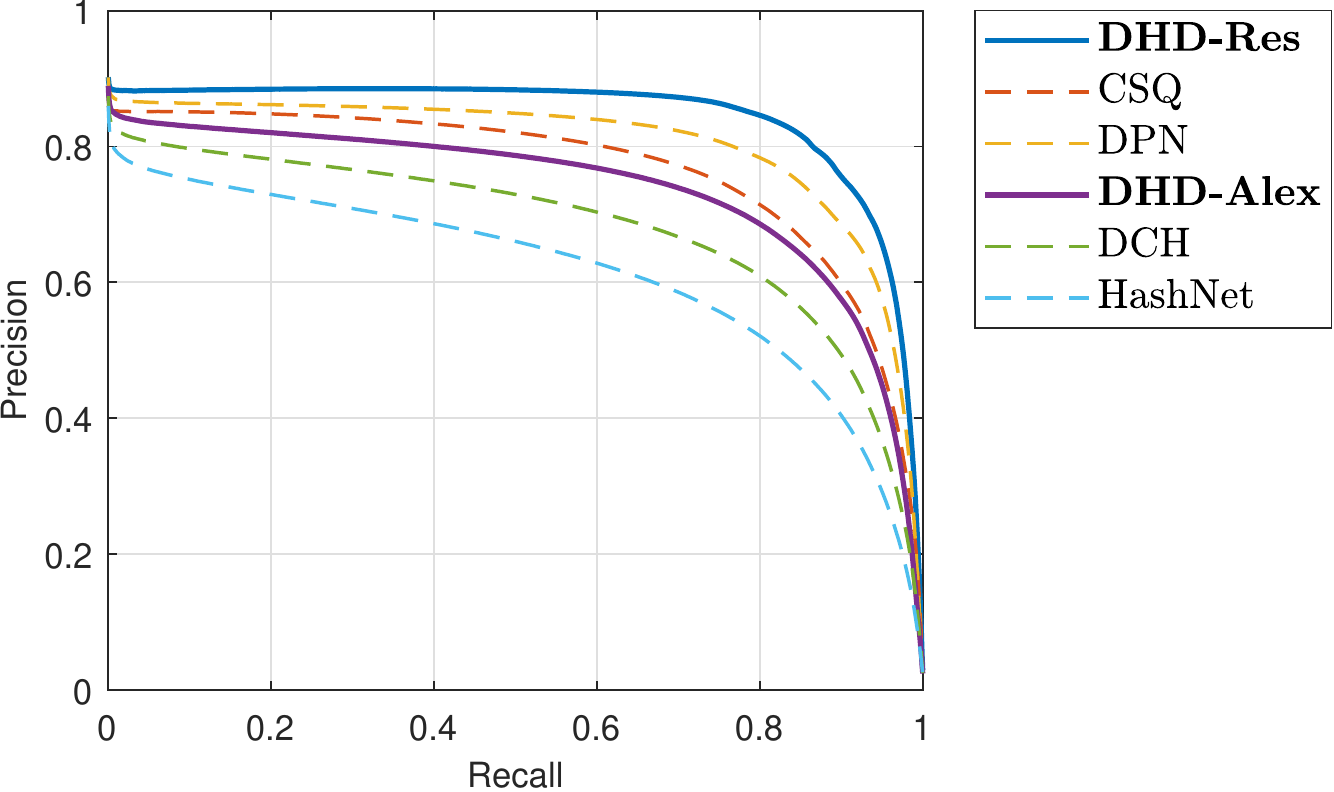}}
\caption{Precision-Recall curves on three image datasets with binary codes @ 64-bits.} 
\label{fig:Figure3}
\end{figure*}

\begin{figure*}[!t]
\centering
\subcaptionbox{ImageNet \label{fig:Figure4_a}}{\includegraphics[height=2.9cm]{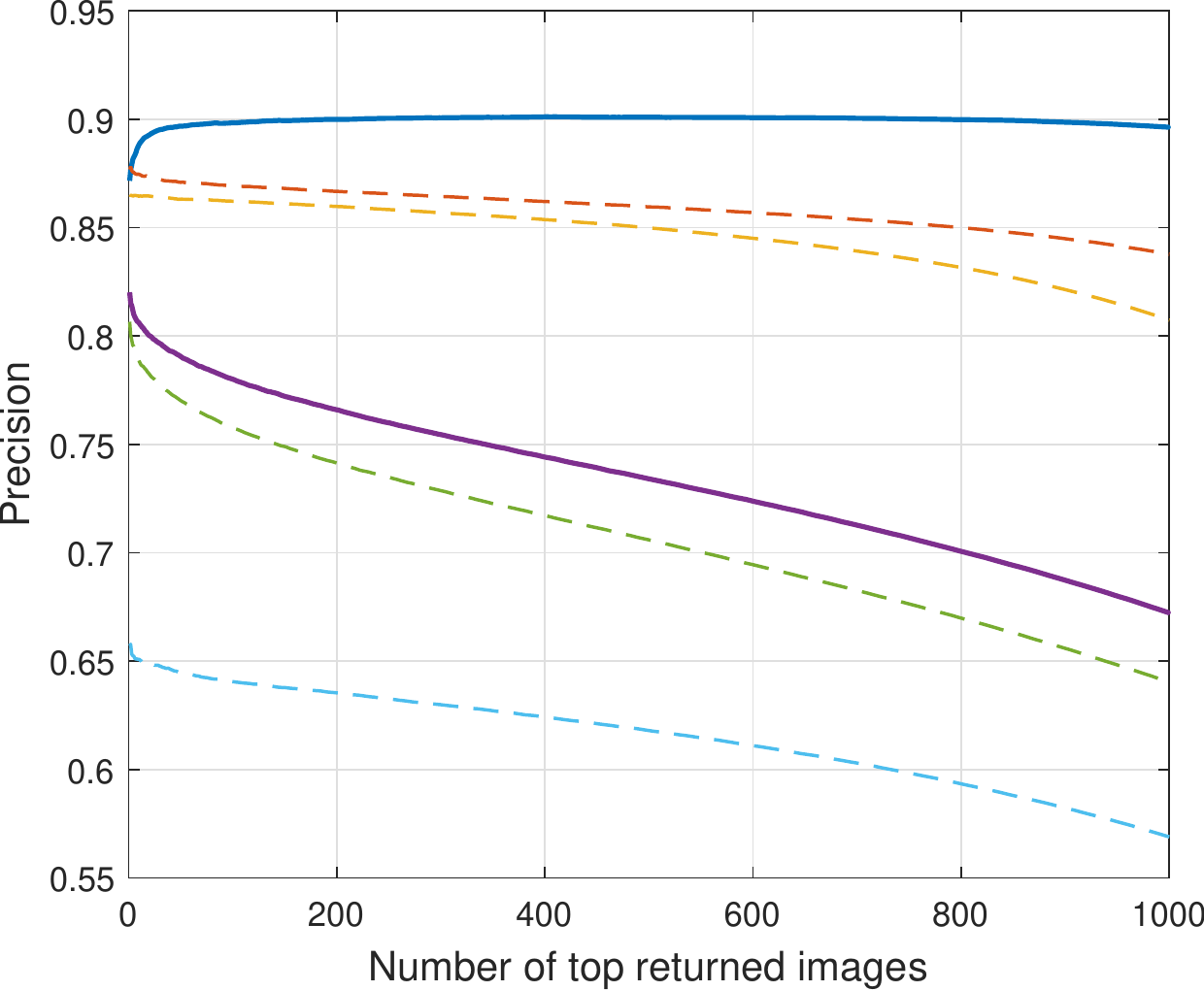}}
\subcaptionbox{NUS-WIDE \label{fig:Figure4_b}}{\includegraphics[height=2.9cm]{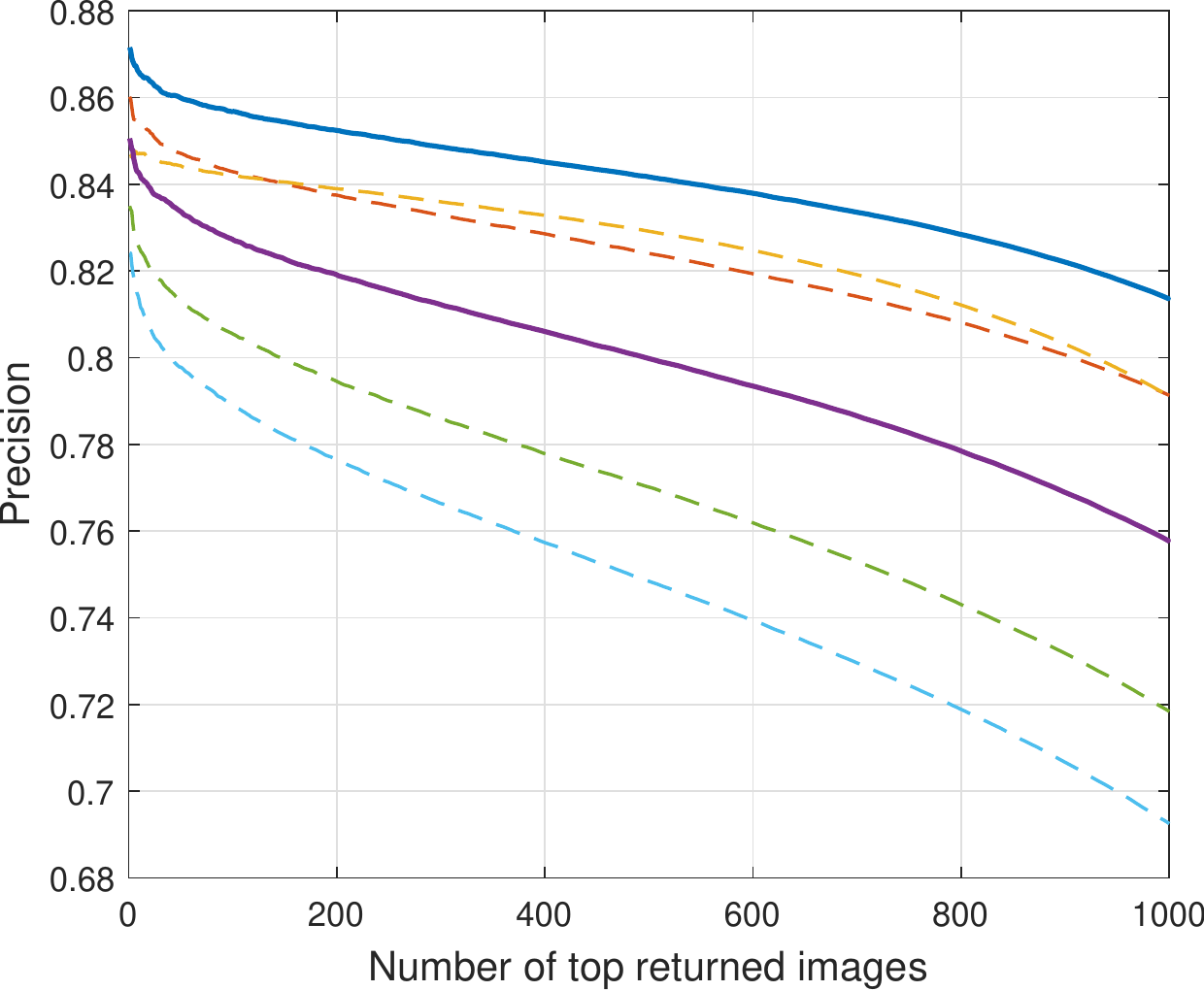}}
\subcaptionbox{MS COCO \qquad\qquad \label{fig:Figure4_c}}{\includegraphics[height=2.9cm]{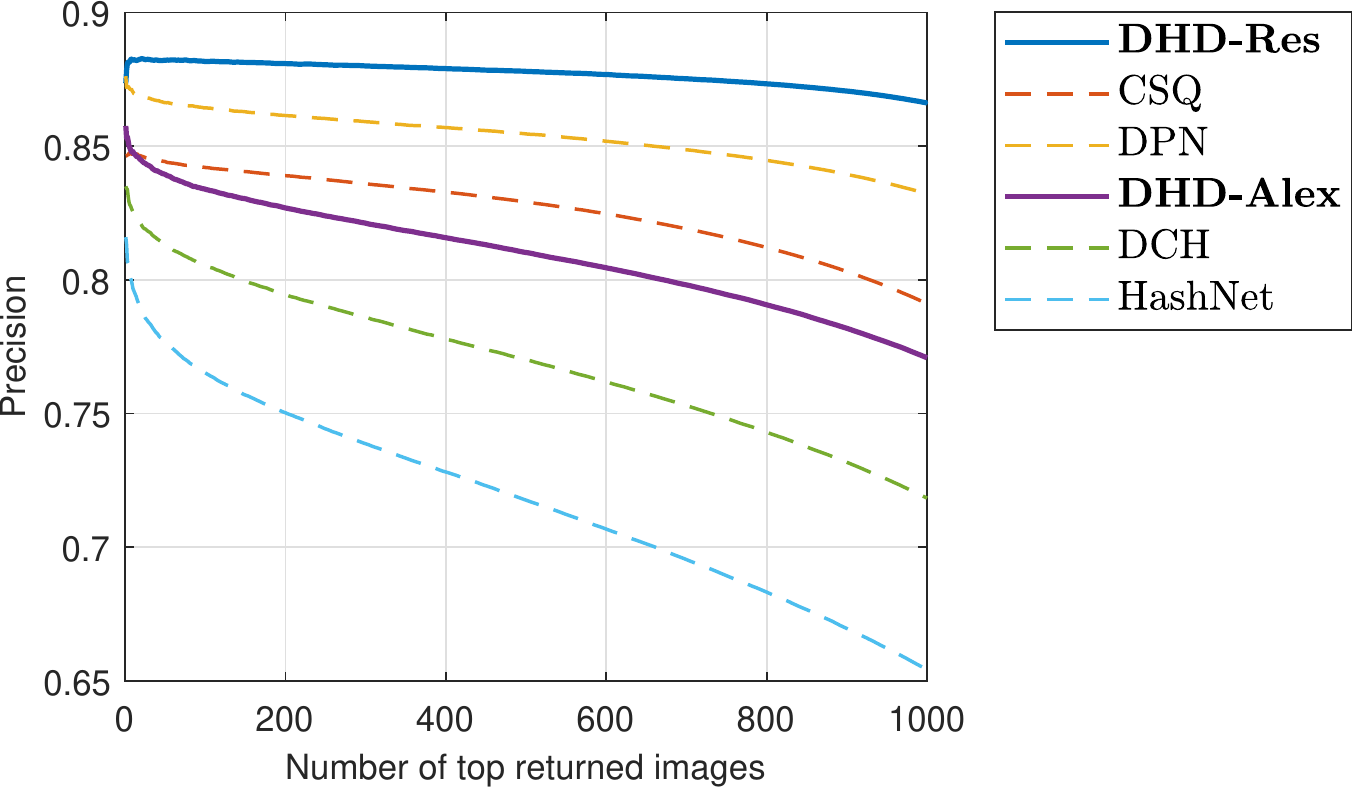}}
\caption{Precision@top-1000 curves on three image datasets with binary codes @ 64-bits.} 
\label{fig:Figure4}
\end{figure*}

To further investigate the retrieval quality of DHD, we deploy the graph of PR curve and precision for the top 1,000 retrieved images at 64 bits. As shown in Figures \ref{fig:Figure3} and \ref{fig:Figure4}, DHD significantly outperforms all the comparison hashing approaches by large margins under these two evaluation metrics. Especially, DHD shows desirable retrieval results in that much higher precision are achieved at lower recall levels, and larger number of top samples are retrieved than all compared methods. These demonstrate the practicality of DHD in real world retrieval cases.

\begin{table*}[!t]
\centering
\caption{mAP scores without ($-$) or with ($+$) Self-distilled Hashing (SdH).}
\begin{adjustbox}{width=0.95\textwidth}
\begin{tabular}{c|c|c|c|c|c|c|c|c|c|c|c|c}
\toprule
\multirow{4}{*}{Method} & \multicolumn{4}{c|}{ImageNet}                                      & \multicolumn{4}{c|}{NUS-WIDE}                                      & \multicolumn{4}{c}{MS COCO}                                       \\ \cmidrule{2-13} 
                        & \multicolumn{2}{c|}{$-$ SdH}    & \multicolumn{2}{c|}{$+$ SdH}  & \multicolumn{2}{c|}{$-$ SdH}    & \multicolumn{2}{c|}{$+$ SdH} & \multicolumn{2}{c|}{$-$ SdH}    & \multicolumn{2}{c}{$+$ SdH} \\ \cmidrule{2-13} 
                        & 16-bit         & 64-bit         & 16-bit         & 64-bit         & 16-bit         & 64-bit         & 16-bit         & 64-bit         & 16-bit         & 64-bit         & 16-bit         & 64-bit         \\ \midrule
HashNet \cite{HashNet}                 & 0.337          & 0.502   & 0.501          & 0.661                       & 0.705          & 0.762     & 0.745          & 0.769               & 0.655          & 0.727       & 0.695          & 0.753      \\
DCH \cite{DCH}                       & 0.571          & 0.597       & 0.640          & 0.673               & 0.748          & 0.767    & 0.754          & 0.771                & 0.669          & 0.697        & 0.703          & 0.746      \\
DPN  \cite{DPN}                          & 0.562          & 0.656   & 0.630          & 0.708                & 0.753          & 0.787    & 0.757          & 0.801           & 0.672          & 0.760           & 0.710          & 0.772        \\
CSQ \cite{CSQ}                            & 0.569          & 0.658      & 0.634          & 0.711         & 0.757          & 0.793         & 0.759          & 0.804         & 0.670          & 0.752             & 0.707          & 0.765        \\ \midrule
$\mathcal{L}_{HP}$                          & 0.574          & 0.660    & 0.642          & 0.715          & 0.759          & 0.798              & 0.766          & 0.812         & 0.706          & 0.759       & 0.725          & 0.775            \\
$\mathcal{L}_{HP}$ + $\mathcal{L}_{bce\text{-}Q}$  & 0.583 & 0.671 & \textbf{0.657} & \textbf{0.721} & 0.775 & 0.806 &\textbf{0.780} & \textbf{0.820} & 0.731 & 0.766 &  \textbf{0.749} & \textbf{0.792} \\
\bottomrule
\end{tabular}
\end{adjustbox}
\label{table:Table3}
\end{table*}

\vspace{0.5em}

\noindent \textbf{Self-distilled Hashing with other methods and ablations.} In order to prove that SdH can be applied to other deep hashing baselines \cite{HashNet,DCH,DPN,CSQ}, we perform retrieval experiments with AlexNet backbone and show the results in Table \ref{table:Table3}. With SdH setup, we employ $\mathcal{T}_T$ and $\mathcal{T}_S$ groups to produce input views, and for without SdH setup, we only use $\mathcal{T}_S$ to generate input views. By comparing Table \ref{table:Table2} and the results without SdH in Table \ref{table:Table3}, we can see that the deep hashing model learned with $\mathcal{T}_S$ is inferior to the model learned with $\mathcal{T}_T$. This is because the use of $\mathcal{T}_S$ increases the chances of emerging discrepancy in representation between Hamming and real space. Otherwise, if the model adopts SdH training to utilize both $\mathcal{T}_T$ and $\mathcal{T}_S$, the retrieval performance can be improved since SdH mitigates discrepancy and properly exploits the power of data augmentation. Intending to see the ablation results of our proposals, we compare the retrieval results between $\mathcal{L}_{HP}$ with the others and find that the trainable setting for the hash centers improve the search quality. Moreover, by combining $\mathcal{L}_{HP}$ with $\mathcal{L}_{bce\text{-}Q}$, the performance gain is obtained for all the bit lengths, showing the power of binary cross entropy-based quantization. Finally, the best mAP scores are achieved when both $\mathcal{L}_{HP}$ and $\mathcal{L}_{bce\text{-}Q}$ are integrated with SdH training, confirming the effectiveness of DHD.

\begin{figure}[!t]
\centering
  \subcaptionbox{ImageNet Hash targets
  \label{fig:Figure5_a}}{\includegraphics[height=4.8cm]{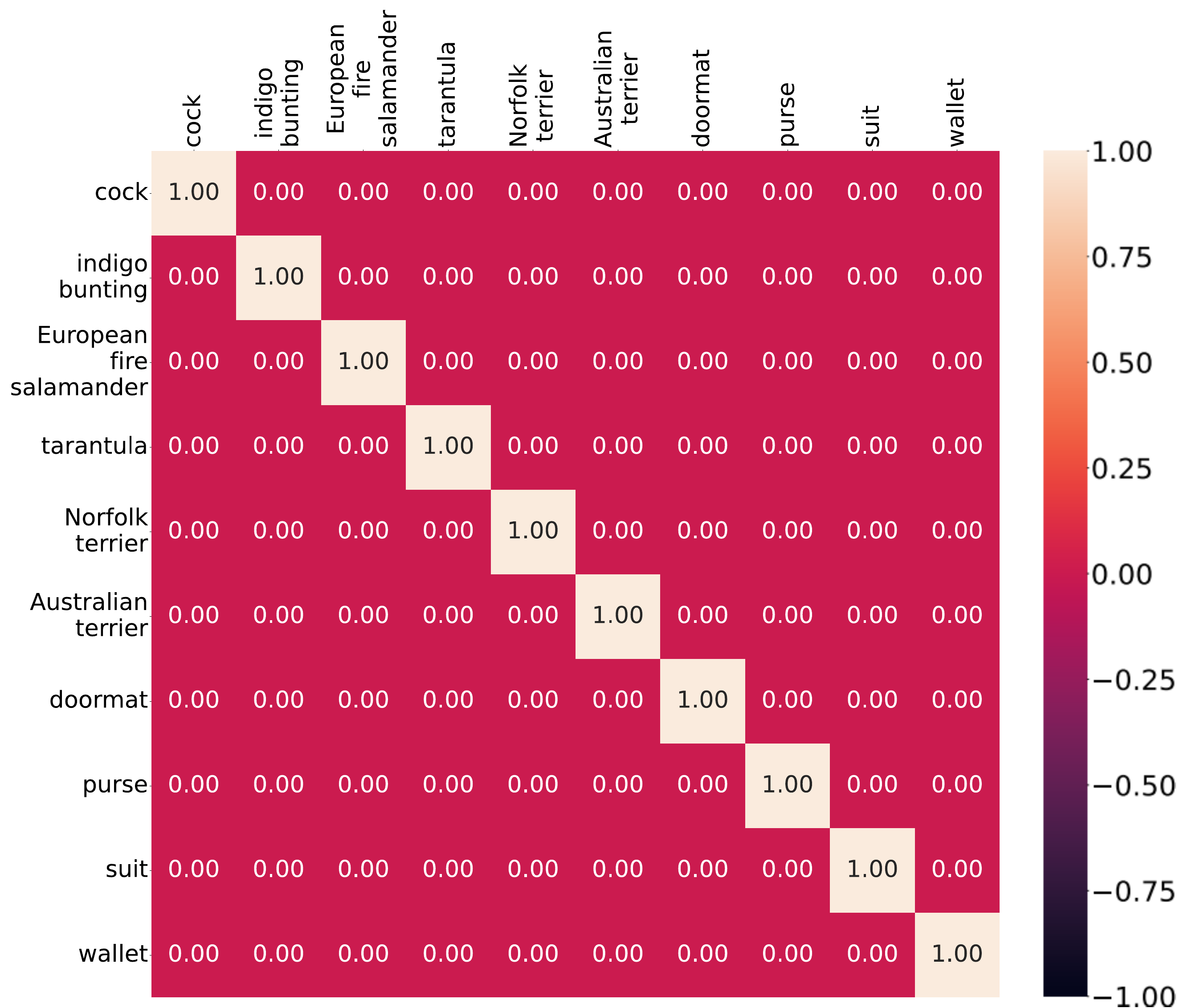}}
  \subcaptionbox{ImageNet Hash proxies \label{fig:Figure5_b}}{\includegraphics[height=4.8cm]{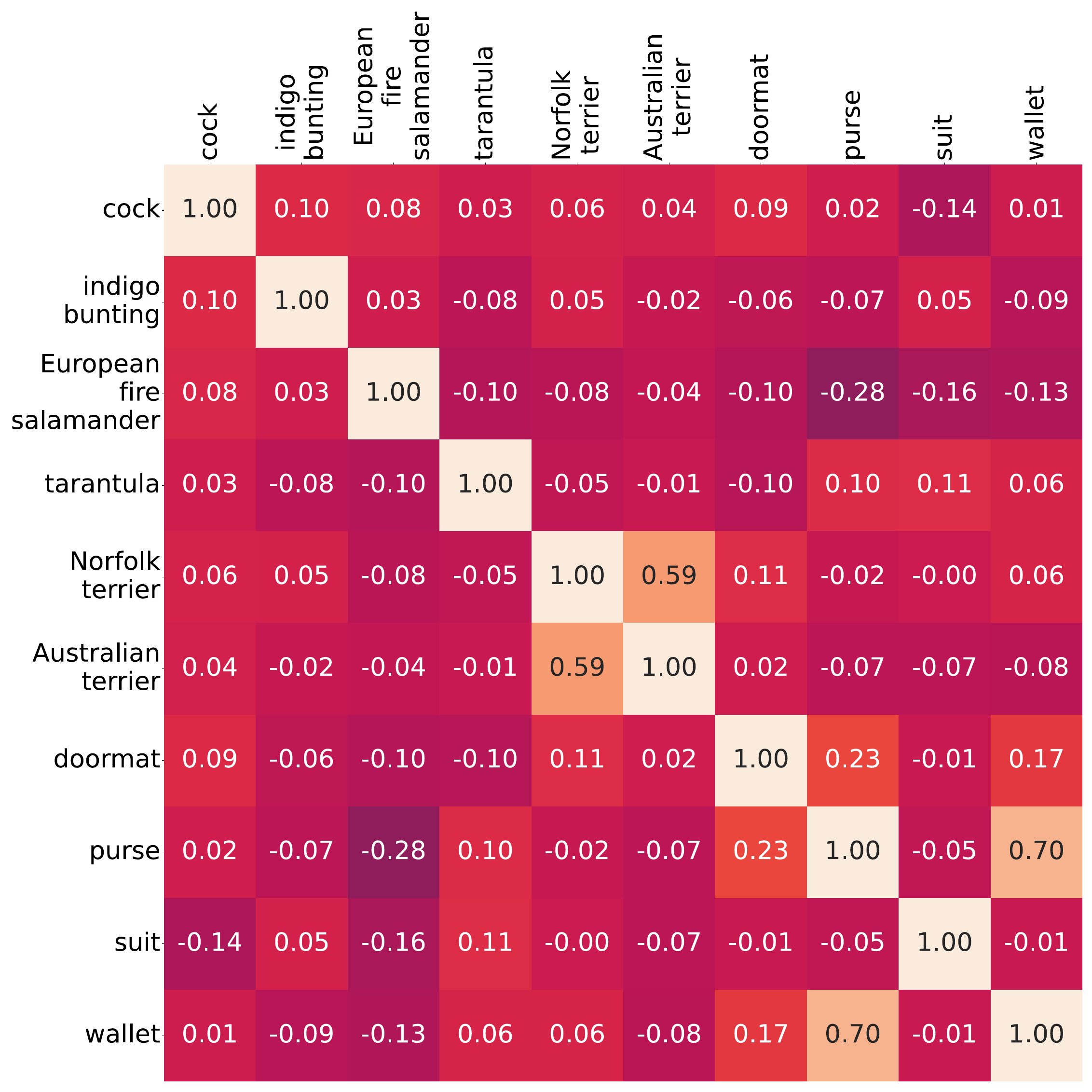}}
  \subcaptionbox{NUS-WIDE Hash Proxies \label{fig:Figure5_c}}{\includegraphics[height=4.8cm]{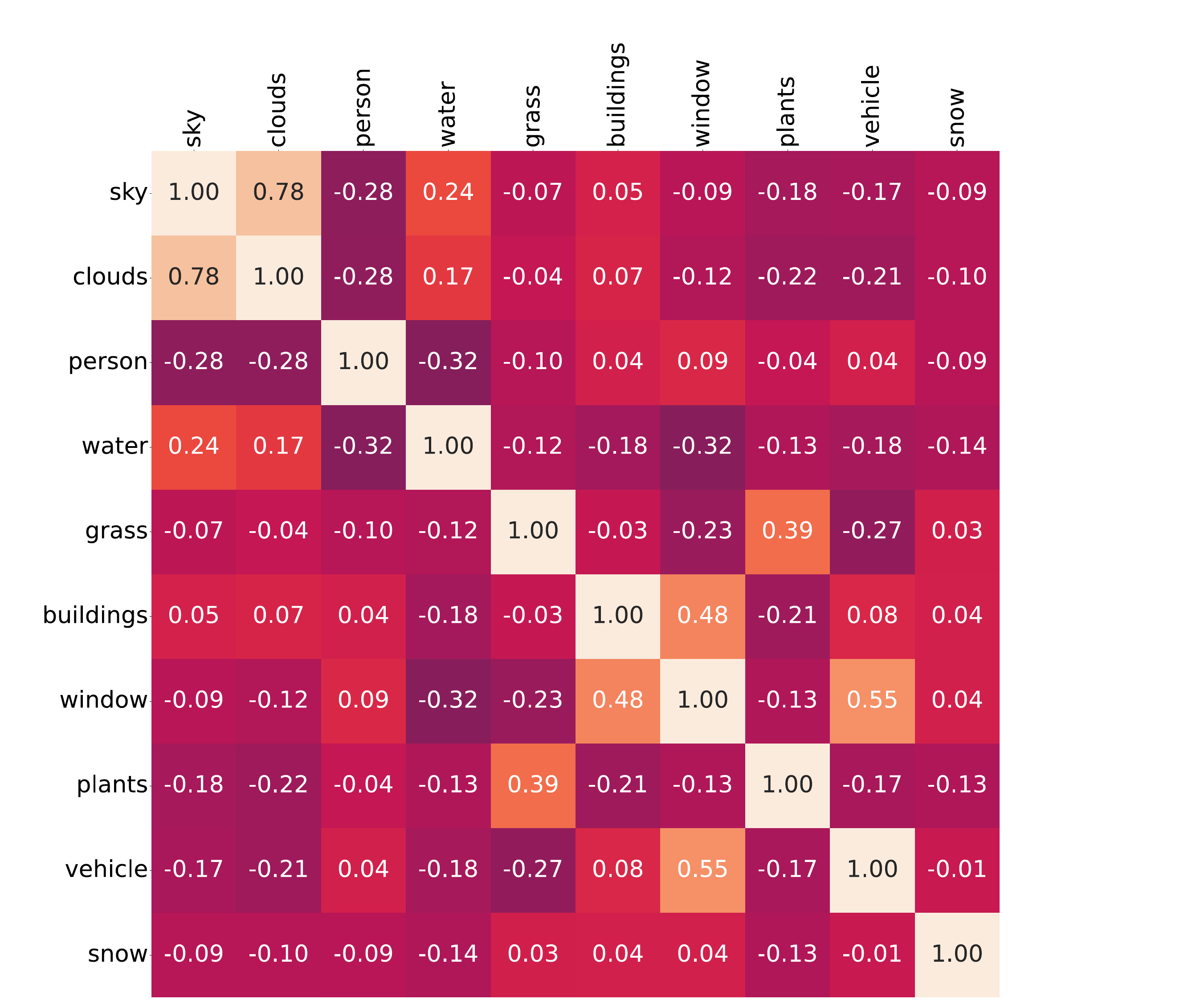}}
  \subcaptionbox{MS COCO Hash Proxies \label{fig:Figure5_d}}{\includegraphics[height=4.8cm]{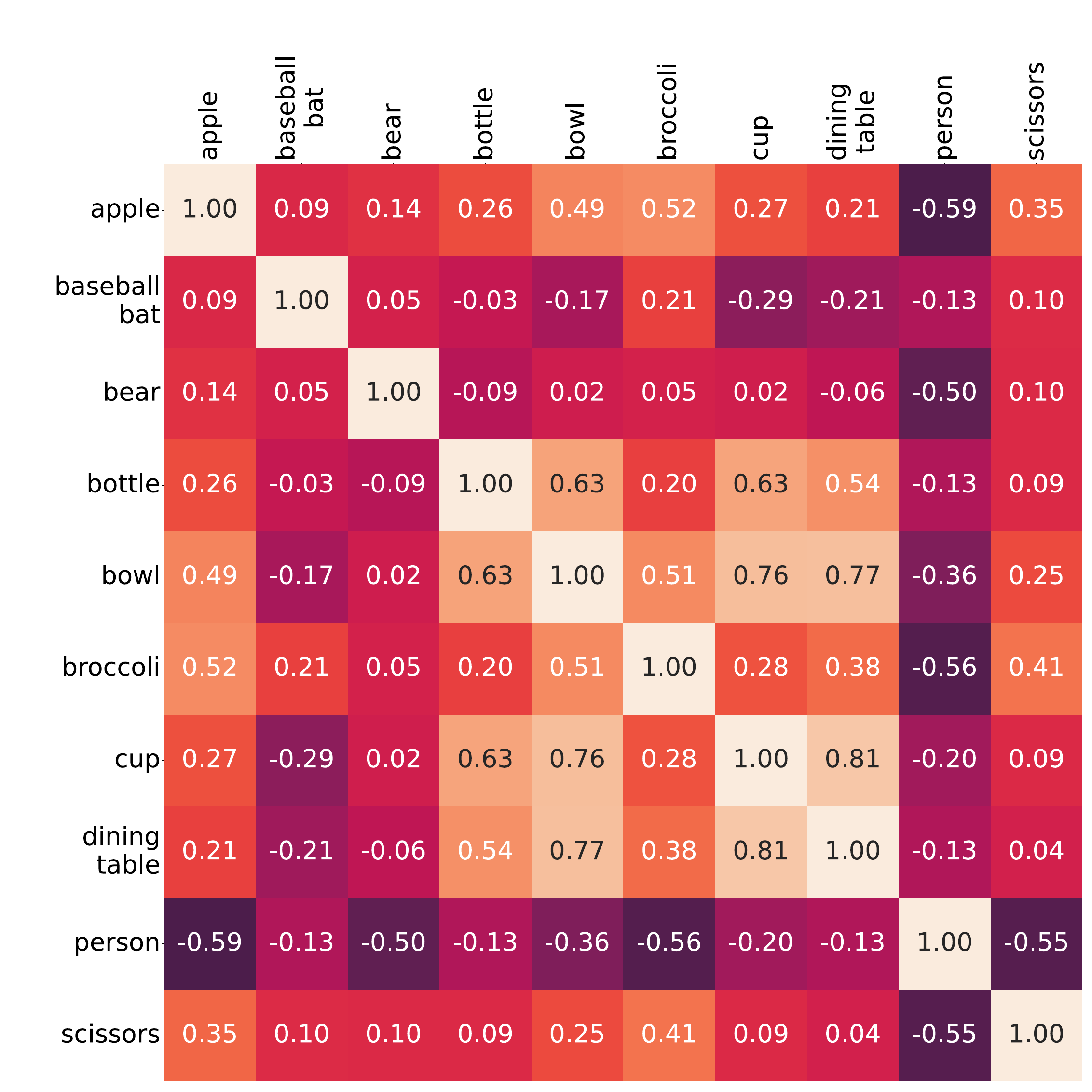}}
\caption{Pairwise cosine similarities to verify the impact of proxy-based hash representation learning. We utilize (a) pre-defined non-trainable hash targets, and (b-d) proposed trainable hash proxies. For simplicity, we show the results of 10 classes selected.}
\label{fig:Figure5}

\end{figure}

\vspace{0.5em}

\noindent \textbf{Trainable Hash Proxies.} In DHD, we employ trainable hash proxies opposed to using predefined orthogonal hash targets \cite{CSQ,DPN,OneLoss}, intending to embed detailed class-wise semantic similarity into the hash representation. We visualize\footnote{Visualized results with all classes for each dataset are shown in the supplementary material.} the pairwise cosine similarities in Figure \ref{fig:Figure5} using ResNet backbone and 64 bit codes to observe the actual alignment between hash proxies. Since hash targets are generated from a Hadamard matrix, they are orthogonal as shown in Figure \ref{fig:Figure5_a}. Therefore, the cosine similarity (Hamming distance) between different hash targets are equal, neglecting the semantic relevance between the hash representations of different classes. Moreover for multi-label cases, label dependencies \cite{Multilabel} are also ignored whether the contents appear simultaneously in an image or not.

On the other hand, trainable hash proxies are designed to embed semantic similarity by themselves during training. Hence, class-wise relevance can be displayed when we compute pairwise cosine similarities as in Figures \ref{fig:Figure5_b} to \ref{fig:Figure5_d}. Specifically for ImageNet, hash proxies of semantically relevant classes have higher similarity, such as \textit{Norfolk terrier-Australian terrier} and \textit{purse-wallet}. Moreover, for multi-label datasets hash proxies of classes that frequently appear together in an image have higher similarity, such as \textit{sky-clouds} and \textit{buildings-window-vehicle} in NUS-WIDE, and \textit{bowl-cup-dining table} in MS COCO. In a nutshell overall, we can confirm that the supervised semantic signals are well guided to represent detailed similarity between the hash proxies, which in turn yields better quality search outcomes.

\begin{figure}[!ht]
\centering
\includegraphics[width=0.72\linewidth]{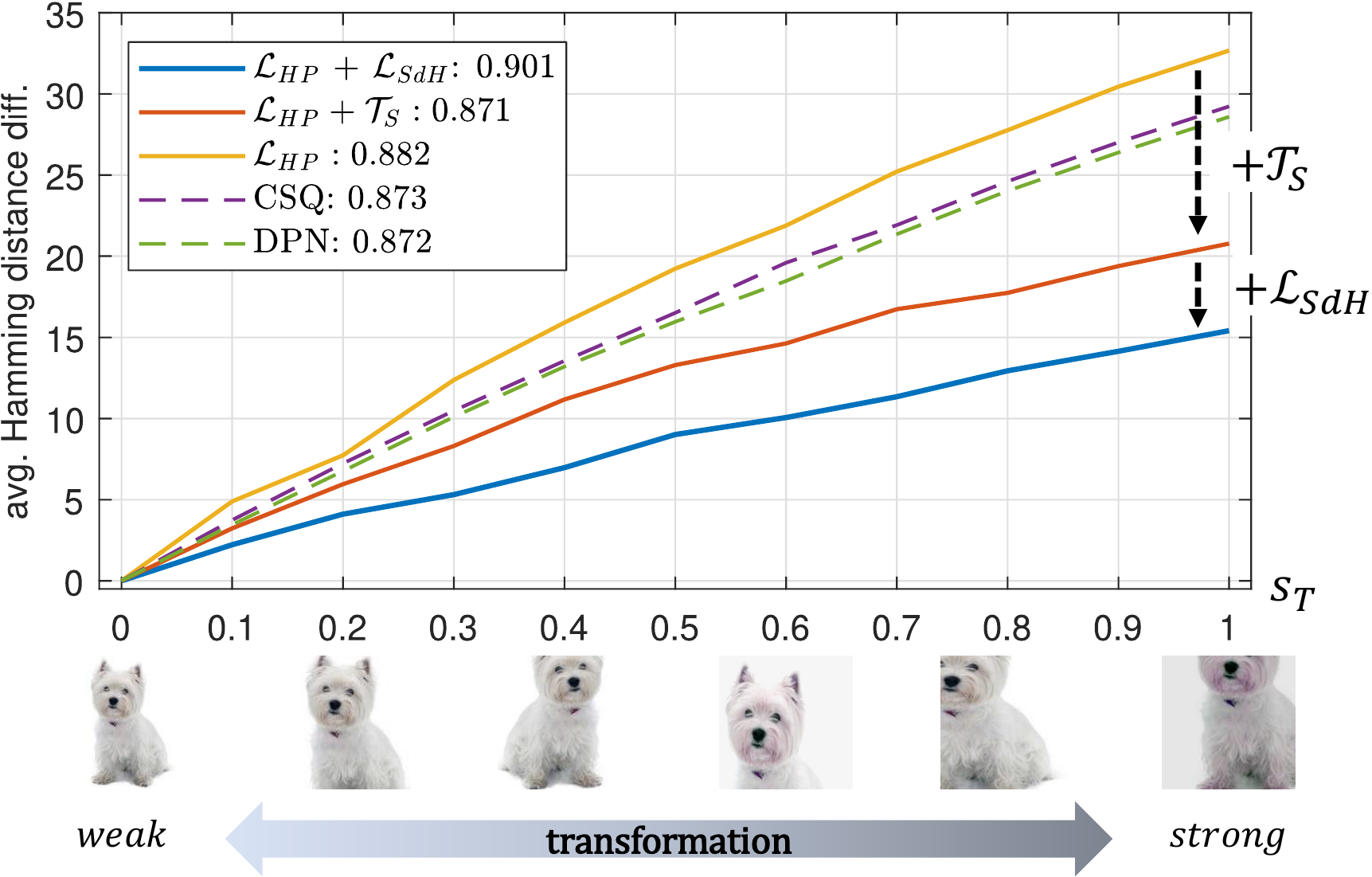}
\caption{Average Hamming distance difference between the original and transformed images of ImageNet query set. By varying the $s_T$, we measure the sensitivity to transformation of ResNet backbone methods, where the numbers in legend indicate mAP. Solid lines present DHD variants, and dotted lines present others. $+\mathcal{T}_S$ denotes strong student augmentation is applied during training. A low slope indicates insensitivity to various transformations, where the blue line (DHD) is the lowest.}
\label{fig:Figure6}
\end{figure}

\vspace{0.5em}

\noindent \textbf{Insensitivity to transformations.} To investigate the sensitivity to transformations, we examine how the binary code shifts when transformed images are fed to ResNet backbone methods, by using ImageNet query set. We measure the average Hamming distance between the untransformed ($s_T=0$) images' output binary codes and the transformed ($s_T$ in (0, 1]) images' output binary codes, as observed in Figure \ref{fig:Figure6}. Here, CSQ \cite{CSQ}, DPN \cite{DPN}, and a model learned with $\mathcal{L}_{HP}$ are trained with weakly-transformed $\mathcal{T}_T$, which in result show sensitivity to transformations due to barely used augmentation. When the augmentation is applied ($\mathcal{L}_{HP} + \mathcal{T}_S$), model is improved to be more robust to transformations, however, the mAP score decreases due to the discrepancy in representation between Hamming and Real space during training. Otherwise, the combined $\mathcal{L}_{HP} + \mathcal{L}_{SdH}$ (blue line) exhibits the highest robustness while achieving the best mAP score, by minimizing discrepancy in representation during training and successfully exploring the potential of strong augmentation.

\begin{table}[!t]
\vspace{-2.0em}
\centering
\caption{mAP scores on unseen deformations.}
\begin{adjustbox}{width=0.47\textwidth}
\begin{tabular}{llc}
\toprule
\multicolumn{1}{c}{Deformation} & \multicolumn{1}{c}{with SdH}  & without SdH\\ \midrule
None                            & \textbf{0.891} \textcolor{teal}{(2.3\% $\uparrow$)}     & 0.871   \\
Cutout                          &\textbf{0.862} \textcolor{teal}{(3.7\% $\uparrow$)}      &   0.827 \\
Dropout                         &\textbf{0.810} \textcolor{teal}{(7.9\% $\uparrow$)}     &   0.765 \\
Zoom in                         & \textbf{0.658} \textcolor{teal}{(19.0\% $\uparrow$)}     & 0.552   \\
Zoom out                        & \textbf{0.816} \textcolor{teal}{(1.4\% $\uparrow$)}       & 0.805 \\
Rotation                        & \textbf{0.856} \textcolor{teal}{(2.4\% $\uparrow$)}       &  0.836\\
Shearing                        & \textbf{0.842} \textcolor{teal}{(2.7\% $\uparrow$)}     & 0.815   \\
Gaussian noise                  & \textbf{0.768} \textcolor{teal}{(10.5\% $\uparrow$)}     & 0.673  \\
\bottomrule
\end{tabular}
\end{adjustbox}
\label{table:Table4}
\vspace{-1.5em}
\end{table}

\vspace{0.5em}

\noindent \textbf{Robustness to unseen deformations.} To further examine the generalization capacity of DHD, we conduct experiments with unseen (not seen during training) transformations \footnote{Detailed deformation setup is listed in the supplementary material.} to inputs following the evaluation protocol utilized in \cite{Deform}. As reported in Table \ref{table:Table4}, deep hashing model with SdH significantly outperforms the model without SdH at all deformations, showing a performance difference of up to 19\% (zoom in). In particular, SdH makes deep hashing model robust to per-pixel deformations such as dropout and Gaussian noise, even though SdH has not included any pixel-level transformations.

\begin{figure}[!ht]
\centering

\centering
  \subcaptionbox{without $\mathcal{L}_{bce\text{-}Q}$ \label{fig:Figure7_a}}{\includegraphics[height=3.7cm]{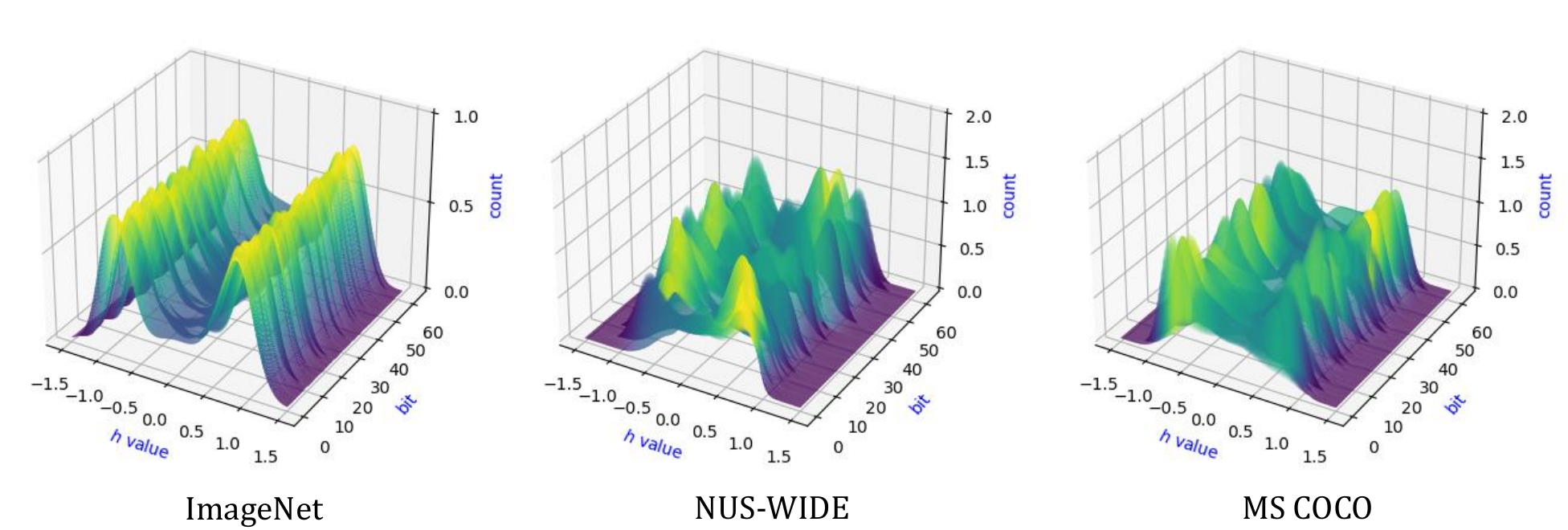}}
  \subcaptionbox{with $\mathcal{L}_{bce\text{-}Q}$ \label{fig:Figure7_b}}{\includegraphics[height=3.7cm]{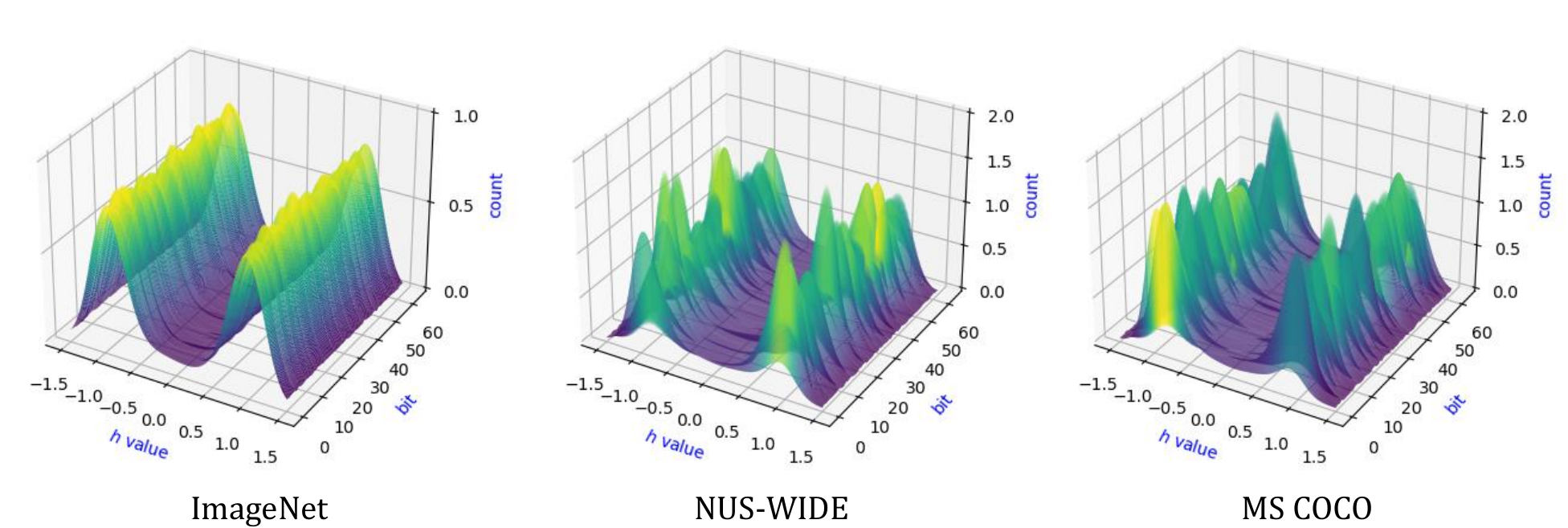}}

\caption{3D visualized histograms to verify the impact of $\mathcal{L}_{bce\text{-}Q}$. $x$-axis presents value of hash element $h$, $y$-axis presents bit position, and $z$-axis presents frequency counts.}
\label{fig:Figure7}
\vspace{-1.8em}
\end{figure}

\vspace{0.5em}

\noindent \textbf{Quantization.} The effect of $\mathcal{L}_{bce\text{-}Q}$ is plotted in Figure \ref{fig:Figure7}. We can see that the binary bits are distributed more evenly and binary-like in \ref{fig:Figure7_b}. This implies that the entropy of bit distribution is much higher when $\mathcal{L}_{bce\text{-}Q}$ is applied, which can show better retrieval accuracy by representing diverse binary codes, as observed and investigated in \cite{BiHalf}.
\vspace{-0.5em}
\section{Conclusion}
\vspace{-0.5em}

In this paper, we proposed a novel Self-distilled Hashing (SdH) scheme which is applicable to deep hashing models during training. By maximizing the cosine similarity between hash codes of different views of an image, SdH minimizes the discrepancy in the representation due to augmentation and leads to increase the robustness of retrieval systems. Additionally, we aimed to embed elaborate semantic similarity into the hash codes with a proxy-based learning, and further impose cross entropy-based quantization loss. With all these proposals, we configured Deep Hash Distillation (DHD) framework that yields the state-of-the-art performance on popular deep hashing benchmarks.
\vspace{-0.5em}
\section{Acknowledgement}
\vspace{-0.5em}

This research was supported in part by NAVER Corp., the National Research Foundation of Korea (NRF) grant
funded by the Korean government (MSIT) (2021R1A2C2007220), and the Institute of Information \& Communications Technology Planning \& Evaluation (IITP) grant funded by the Korean government (MSIT) [NO.2021-0-01343, Artificial Intelligence Graduate School Program (Seoul National University)].
\clearpage
%
%
\bibliographystyle{splncs04}
\bibliography{egbib}
\end{document}